\documentclass[sigconf]{acmart}
\AtBeginDocument{%
  \providecommand\BibTeX{{%
    \normalfont B\kern-0.5em{\scshape i\kern-0.25em b}\kern-0.8em\TeX}}}

\setcopyright{acmlicensed}
\copyrightyear{2024}
\acmYear{2024}
\acmDOI{XXXXXXX.XXXXXXX}

\acmConference[KDD '24]{ACM KDD 2024}{August 24--29, 2024}{Barcelona, Spain}
\acmISBN{978-1-4503-XXXX-X/18/06}

\begin{document}

%%
%% The "title" command has an optional parameter,
%% allowing the author to define a "short title" to be used in page headers.
\title{AI Safety in Practice: Enhancing Adversarial Robustness in Multimodal Image Captioning}

%%
%% The "author" command and its associated commands are used to define
%% the authors and their affiliations.
%% Of note is the shared affiliation of the first two authors, and the
%% "authornote" and "authornotemark" commands
%% used to denote shared contribution to the research.
% \author{Blind Authors}
% \affiliation{%
%   \institution{Blind University}
%   \streetaddress{Blind Address}
%   \city{Blind City}
%   \country{Blind State}}
% \email{blind@lab.ai}

\author{Maisha Binte Rashid}
\affiliation{%
  \institution{Baylor University}
  \streetaddress{One Bear Place}
  \city{Waco}
  \country{Texas}}
\email{maisha\_rashid1@baylor.edu}

\author{Pablo Rivas}
\affiliation{%
  \institution{Baylor University}
  \streetaddress{One Bear Place}
  \city{Waco}
  \country{Texas}}
\email{pablo\_rivas@baylor.edu}

%%
%% By default, the full list of authors will be used in the page
%% headers. Often, this list is too long, and will overlap
%% other information printed in the page headers. This command allows
%% the author to define a more concise list
%% of authors' names for this purpose.
% \renewcommand{\shortauthors}{Blind Authors}
\renewcommand{\shortauthors}{Maisha Binte Rashid and Pablo Rivas}

%%
%% The abstract is a short summary of the work to be presented in the
%% article.
\begin{abstract}
  Multimodal machine learning models that combine visual and textual data are increasingly being deployed in critical applications, raising significant safety and security concerns due to their vulnerability to adversarial attacks. This paper presents an effective strategy to enhance the robustness of multimodal image captioning models against such attacks. By leveraging the Fast Gradient Sign Method (FGSM) to generate adversarial examples and incorporating adversarial training techniques, we demonstrate improved model robustness on two benchmark datasets: Flickr8k and COCO. Our findings indicate that selectively training only the text decoder of the multimodal architecture shows performance comparable to full adversarial training while offering increased computational efficiency. This targeted approach suggests a balance between robustness and training costs, facilitating the ethical deployment of multimodal AI systems across various domains. 
\end{abstract}

%%
%% The code below is generated by the tool at http://dl.acm.org/ccs.cfm.
%% Please copy and paste the code instead of the example below.
%%


%%
%% Keywords. The author(s) should pick words that accurately describe
%% the work being presented. Separate the keywords with commas.

%% A "teaser" image appears between the author and affiliation
%% information and the body of the document, and typically spans the
%% page.
% \begin{teaserfigure}
%   \includegraphics[width=\textwidth]{sampleteaser}
%   \caption{Seattle Mariners at Spring Training, 2010.}
%   \Description{Enjoying the baseball game from the third-base
%   seats. Ichiro Suzuki preparing to bat.}
%   \label{fig:teaser}
% \end{teaserfigure}

% \received{20 February 2007}
% \received[revised]{12 March 2009}
% \received[accepted]{5 June 2009}

%%
%% This command processes the author and affiliation and title
%% information and builds the first part of the formatted document.
\maketitle

\section{Introduction}
Multimodal machine learning is an advanced area of artificial intelligence that integrates and processes multiple types of data inputs, such as text, images, and audio, to perform tasks that mimic human cognitive abilities~\cite{gao2020survey}. This integration allows the model to leverage the strengths of each modality, leading to richer interpretations and more accurate predictions that could be achieved by models processing the data in isolation. One of the most compelling applications of multimodal learning involves the combination of visual and textual data, commonly referred to as image-text pairs. This pairing is particularly significant as it mirrors the way humans often receive and interpret information, making the study of these models not only interesting but also aligned with natural human communication patterns.

The increasing deployment of multimodal models in critical applications also raises significant safety and security concerns. These models, like all machine learning systems, are susceptible to adversarial attacks which means intentional inputs designed to confuse the model and provoke incorrect outputs~\cite{ozdag2018adversarial, evtimov2020adversarial}. The robustness of multimodal models, therefore, becomes a critical area of focus. Ensuring these models can withstand adversarial attacks is not just a technical necessity but a safety imperative, particularly when these models are employed in sensitive contexts such as autonomous driving, healthcare, and content moderation. 

Making multimodal machine learning models more robust against adversarial attacks is a key challenge~\cite{schlarmann2023adversarial}. These models, which combine different types of data like text and images, can be especially vulnerable because of the way these data types interact. Traditional methods that focus on making each type of data robust on its own can be very demanding and might not address all the issues.

In this paper, we look at a simpler and more efficient approach. Instead of trying to make the entire model robust, we focus on improving just one part of it. We apply adversarial training techniques specifically to the text decoder part of our image captioning model. Our experiments with the Flickr8k and COCO datasets show that this method works well. Training only the text decoder, while keeping the image encoder fixed, gives us results almost as good as training the whole model. However, when we fix the text decoder and train only the image encoder, the performance drops significantly. This shows that the text decoder plays a crucial role in making the model robust against attacks.

This paper helps improve ethical AI by showing a focused way to make multimodal models stronger. This method not only improves the technical side of AI but also helps build public trust in AI systems used in different areas.

This paper is organized as follows: Section 2 reviews related work on adversarial attacks and defenses in multimodal machine learning. Section 3 describes our model architecture and the adversarial training method. Section 4 presents the experimental setup and results on the Flickr8k and COCO datasets. Section 5 discusses what our findings mean for developing robust and ethical AI systems. Finally, Section 6 concludes the paper and suggests future research directions.

\section{Background}
In~\cite{yu2020investigating}, the authors investigated the vulnerability of multimodal deep learning models to adversarial attacks, finding that even if only one input modality is attacked, the overall performance degrades significantly. The authors highlight the need to explore ways to obtain robust features from multimodal data to achieve useful information from each modality and improve adversarial defense mechanism. To develop adversarial defense mechanisms we explored different adversarial attacks in machine learning models. 
One of the first and best-known adversarial attack techniques is the Fast Gradient Sign Method (FGSM), which involves applying perturbations to the input data in the direction of the gradient of the loss function with respect to the input data to produce adversarial examples~\cite{goodfellow2014explaining}. Another adversarial attack similar to FGSM, which creates the attack iteratively, is Projected Gradient Descent (PGD)~\cite{madry2017towards}. PGD generates adversarial examples by iteratively taking small steps toward the direction of maximum loss function and projecting the outcome back onto the allowed data range at each step. Other well-known adversarial attack strategies are Jacobian Saliency Map Adversary (JSMA) and C\&W Attack.
JSMA is an iterative adversarial attack technique that computes a saliency map based on the Jacobian matrix to identify the most significant pixels to perturb in order to mislead neural network classifiers~\cite{papernot2016limitations}. The C\&W attack, proposed by Carlini and Wagner, is an powerful and efficient optimization-based adversarial attack that finds quasi-imperceptible perturbations to input examples that reliably cause misclassification in neural networks~\cite{carlini2017towards}.

The authors in~\cite{noever2021reading} discussed about adversarial attack in multimodal machine learning model like CLIP in their paper. They presented new types of adversarial attacks against the multi-modal neural network model CLIP, which integrates the detection of visual objects with text reading. Basic typographical manipulations like misspellings and font changes are included in the attacks, along with conceptual ones that use contradicting text and image inputs to trick the model. Adversarial training is one of the most common defense mechanism for adversarial attack in machine learning models. Adversarial training is using adversarial examples as samples in the training phase of the model. The authors in~\cite{shafahi2019adversarial} proposed a fast adversarial training algorithm that produces robust models at a low computational cost by recycling the gradient information computed during training. The method is able to train robust image classifiers for the ImageNet dataset in a short time on a modest hardware setup.
\section{Methodology}
\subsection{Model Architecture}
This work uses an architecture for the image captioning task that combines GPT-2 and Vision Transformer (ViT) models~\cite{dosovitskiy2020image}, with the GPT-2~\cite{radford2019language} acting as a decoder and the ViT as an encoder. The ViT model divides an input image of 224 x 224 pixels into 16 x 16 pixel patches. Then, these patches are flattened and projected linearly into a space with dimensions of 768. The transformer encoder receives the image patch sequence and a [CLS] token. The encoder consists of multiple layers of position-wise feedforward and multi-headed self-attention networks. The \texttt{[CLS]} token, which compiles data from all patches, represents the whole image. This token's final hidden state at the ViT encoder's output captures the contextualized global aspects of the image, which are essential for generating a caption.  

The decoder obtains the encoded image features from the ViT's [CLS] token, which uses them as the starting point for caption generation. Through many layers of masked multi-headed self-attention, the GPT-2 model processes this input, allowing each point in the output sequence to attend only to earlier positions in the sequence. This arrangement preserves the autoregressive characteristic by ensuring that each word in the caption is generated solely based on the words that came before it. The GPT-2 model produces a sequence of tokens that comprise the image's caption. Every token is created one after the other, and the model predicts the subsequent token by analyzing the preceding tokens and the contextual data extracted from the image.  

\subsection{Adversarial Attack}
In our work, we used the adversarial example technique to create adversarial attacks. Specifically, we employed the Fast Gradient Sign Method (FGSM) to generate these examples. FGSM is a method for crafting adversarial examples, which are inputs to machine learning models that have been intentionally modified to cause the model to make incorrect predictions~\cite{goodfellow2014explaining}.

Given an input $\mathbf{x}$ and its true label $y$, we define the loss function $J(\theta, \mathbf{x}, y)$, where $\theta$ represents the parameters of the model. The goal is to find a small perturbation $\mathbf{\eta}$ such that the perturbed input $\mathbf{x}' = \mathbf{x} + \mathbf{\eta}$ leads to a misclassification. FGSM achieves this by using the gradient of the loss function with respect to the input $\mathbf{x}$.

First, we compute the gradient of the loss function:
\[
\mathbf{g} = \nabla_{\mathbf{x}} J(\theta, \mathbf{x}, y).
\]

Then, the adversarial example $\mathbf{x}'$ is generated by adding a perturbation in the direction of the sign of this gradient:
\[
\mathbf{x}' = \mathbf{x} + \epsilon \cdot \text{sign}(\mathbf{g}).
\]

Here, $\epsilon$ is a small constant that controls the magnitude of the perturbation. The sign function, $\text{sign}(\mathbf{g})$, is applied element-wise to the gradient, ensuring that each element of the perturbation has the same magnitude, $\epsilon$.

\begin{table*}[h!]
\centering
\caption{Examples of adversarial perturbations from the Flickr dataset (top three rows) and COCO (bottom two rows).}\label{tbl:pert}
\begin{tabular}{|c|c|c|}
\hline
\textbf{Original} & \textbf{Perturbed Example} & \textbf{Difference} \\ \hline
\includegraphics[width=0.22\linewidth]{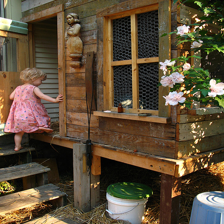} & \includegraphics[width=0.22\linewidth]{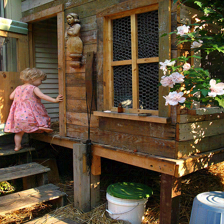} & \includegraphics[width=0.22\linewidth]{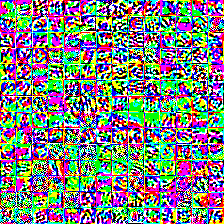} \\ \hline
\includegraphics[width=0.22\linewidth]{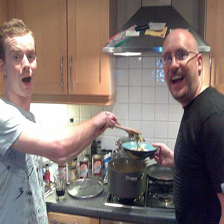} & \includegraphics[width=0.22\linewidth]{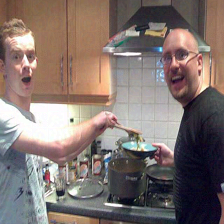} & \includegraphics[width=0.22\linewidth]{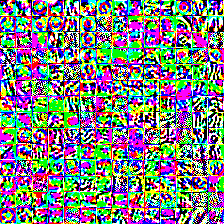} \\ \hline
\includegraphics[width=0.22\linewidth]{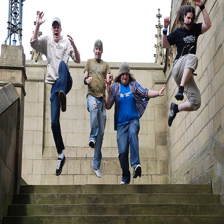} & \includegraphics[width=0.22\linewidth]{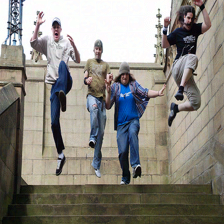} & \includegraphics[width=0.22\linewidth]{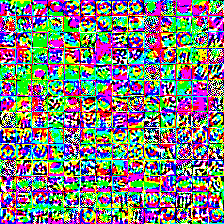} \\ \hline
\includegraphics[width=0.22\linewidth]{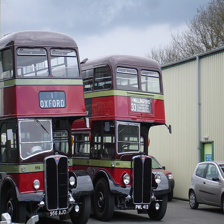} & \includegraphics[width=0.22\linewidth]{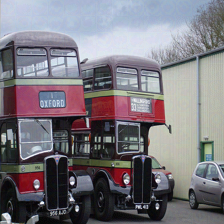} & \includegraphics[width=0.22\linewidth]{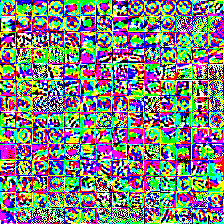} \\ \hline
\includegraphics[width=0.22\linewidth]{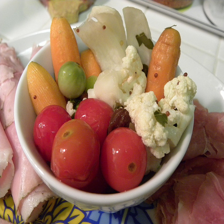} & \includegraphics[width=0.22\linewidth]{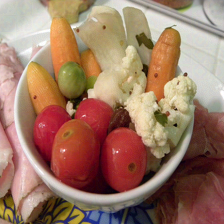} & \includegraphics[width=0.22\linewidth]{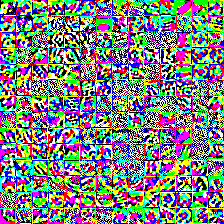} \\ \hline
\end{tabular}
\end{table*}

Table~\ref{tbl:pert} shows examples of the adversarial perturbations generated using FGSM on the Flickr8k and COCO datasets. Despite its simplicity, FGSM is effective in generating adversarial examples that are close to the original inputs but cause the model to make incorrect predictions. This method is computationally efficient and serves as a useful tool for testing and improving the robustness of machine learning models. This table shows how adversarial attacks are imperceptible to human eyes, as shown in the second column. The third column demonstrates the perturbations by showing the difference between the original and perturbed images. In the 'Difference' column, we can see square or patch-like patterns. This happens because of ViT's architecture, which processes input images by dividing them into patches. FGSM leverages the gradients computed for each patch, generating perturbations that exploit the vulnerabilities within these regions. Table \ref{tbl:pert} underscores the necessity for robust models capable of defying such attacks. By incorporating adversarial examples into the training process, we can enhance the model's ability to defend against such attacks and improve its overall robustness.

\subsection{Dataset}
In this study, we leverage two widely-used and benchmark datasets for the image captioning task: Flickr8k and COCO (Common Objects in Context).
\subsubsection{Flickr8k Dataset}
The Flickr8k dataset comprises 8,000 images, each accompanied by five human-annotated captions. The images are sourced from Flickr, a popular online photo-sharing platform, and are categorized into 20 distinct classes.
\subsubsection{COCO Dataset}
The COCO dataset is a larger and more challenging resource for image captioning, containing 123,287 images with five captions per image. Unlike the Flickr8k dataset, the images in COCO are sourced from various online platforms, including Flickr, and are categorized into 80 classes. 
\subsection{Evaluation Metric}
To evaluate the performance for image captioning task, we used Bilingual Evaluation Understudy (BLEU) score~\cite{papineni2002bleu}. Comparing the generated captions to the reference captions in the dataset generates a similarity score known as the BLEU score. Higher scores correspond to a closer match. The BLEU score computes the n-gram overlap between the model's output and the ground truth. There is a perfect match between the generated and reference captions when the BLEU score is 1, from 0 to 1. 
\subsection{Experiment}
We used a multi-phased approach incorporating adversarial training to increase the robustness of our multimodal image captioning model, which combines a Vision Transformer (ViT) encoder and a GPT-2 decoder. We conducted our experiment in five phases as follows:
\begin{enumerate}
    \item We trained the ViT-GPT-2 architecture on the Flickr8k dataset without considering adversarial factors, creating a baseline model that served as a benchmark for assessing the following experiments.
    \item We applied the FGSM to create adversarial examples for the training and test sets. We used these samples to train the model and evaluated the performance of the model. 
    \item To improve robustness, we performed adversarial training on the model using both the original data and adversarial examples.
    \item We investigated a more focused strategy to further improve the model's robustness. Initially, we used only the GPT-2 decoder for adversarial training while freezing the ViT encoder.
    \item We reversed the process by freezing the GPT-2 decoder and using the ViT encoder alone for adversarial training.
\end{enumerate}

We conducted steps four and five three times each and calculated the average of these trials.  
Focusing on a single modality at a time allowed us to implement more effective adversarial defense mechanisms without having to scale them across the entire model. We repeated these five steps of experiment with COCO dataset as well to access the robustness of our model. For both the datasets we have used the same perturbation magnitude, $\epsilon$ = 0.1.

\section{Results}
After conducting experiments on both the Flickr8k and COCO datasets, we evaluated the performance of our multimodal machine learning models using the BLEU score metric. The goal of these experiments was to find an efficient way to increase the robustness of these models against adversarial attacks. As shown in Table \ref{tab:model_flickr}, our baseline model, trained on the original data, initially achieved the best performance. However, when we generated adversarial examples using the FGSM and trained the model with these samples, its performance decreased significantly. This suggests that our multimodal model is not robust against adversarial attacks, as generating a single adversarial example for each image in the dataset was sufficient to degrade its performance.

To enhance the model's robustness, we trained it with both the original data and the adversarial examples. This adversarial training approach improved the model's performance, indicating that incorporating adversarial examples during training can help mitigate the impact of such attacks.

Aiming to find an efficient way to increase the robustness of our multimodal model, we investigated whether training only one of its components (either the image model or the text model) would be sufficient. First, we froze the image model (ViT) and trained the text model with both the original data and the adversarial examples. As shown in Table \ref{tab:model_flickr}, the model's performance in this case was lower than with full adversarial training, although the difference was not significant. Alternatively, we froze the text model (GPT) and trained the image model; in this case, we can see the performance is much lower than the full adversarial training's performance. 
\begin{table}[h]
\centering
\caption{Model performance on training and testing datasets using Flickr8k dataset.}
\begin{tabular}{lrr}
\hline
Model       & Train Set & Test Set \\ \hline
Baseline BLEU Score    & 0.285 & 0.232 \\ 
Adversarial Example & 0.164 & 0.143  \\
Adversarial Training & 0.217 & 0.215  \\
Adversarial Training by freezing ViT  & 0.200 & 0.181 \\
Adversarial Training by freezing GPT  & 0.060 & 0.050 \\ \hline
\end{tabular}
\label{tab:model_flickr}
\end{table}

We conducted similar experiments using the COCO dataset, and the results are presented in Table \ref{tab:model_coco}. The baseline performance on the COCO dataset was higher than on Flickr8k. However, when the model was trained solely on adversarial examples generated from the COCO dataset, the performance decreased significantly, as expected. In the third phase, where we trained the model on both the original data and the adversarial samples, the result was comparable to the baseline model's performance.

When we froze the image model (ViT) and trained only the text model, the difference in BLEU score compared to the full adversarial training approach was relatively small. This observation aligns with our findings from the Flickr8k dataset.
\begin{table}[h]
\centering
\caption{Model performance on training and testing datasets using COCO dataset.}
\begin{tabular}{lrr}
\hline
Model       & Train Set & Test Set \\ \hline
Baseline BLEU Score    & 0.3000 & 0.2812 \\ 
Adversarial Example & 0.1464 & 0.1430  \\
Adversarial Training & 0.2910 & 0.2890  \\
Adversarial Training by freezing ViT  & 0.2601 & 0.2500 \\
Adversarial Training by freezing GPT  & 0.0900 & 0.0920 \\ \hline
\end{tabular}
\label{tab:model_coco}
\end{table}

\section{Discussion}
Through experiments on both the Flickr8k and COCO datasets, we can conclude that while the adversarial training technique improves the robustness of our multimodal models against adversarial attacks, their performance does not fully match the baseline levels achieved with clean data. However, the adversarially trained models showed performance close to the baseline, suggesting that this approach can effectively mitigate the impact of adversarial attacks.

Our findings suggest that performing adversarial training only in the text model can achieve comparable performance to full adversarial training. However, freezing the text model or the GPT-2 model during training significantly degrades performance. When the ViT encoder is frozen and adversarial training is performed solely on the GPT-2 decoder, the model still benefits from the ViT encoder's robust feature extraction capabilities. The ViT encoder continues to deliver consistent and reliable image features, enabling the GPT-2 decoder to concentrate on learning to generate text robustly. In contrast, when the GPT-2 decoder is frozen and adversarial training is applied solely to the ViT encoder, the model's capacity to generate coherent and contextually relevant text is hampered. Since the parameters of the GPT-2 decoder remain fixed, it cannot adjust to the alterations in the image features produced by the ViT encoder during adversarial training. Consequently, this lack of adaptability in the text generation process results in a decline in performance. In conclusion, this finding presents an opportunity to increase computational efficiency by focusing the training efforts on a text modality, with a modest trade-off in terms of performance compared to training the entire multimodal model.

\section{Conclusion}
Our research has shown important steps forward in making multimodal machine learning models more robust, especially for tasks like image captioning. We combined Vision Transformer (ViT) and GPT-2 architectures and used adversarial training methods to focus on specific parts of the model. This focused approach helps protect against adversarial attacks without needing to train the entire model, which saves time and resources. Through our experiments with the Flickr8k and COCO datasets, we found that applying adversarial training to just the text decoder can make the models much more robust. While the performance of these models with adversarial training is slightly lower than those trained only on clean data, the trade-off is minimal. This means we can improve model safety without losing much accuracy. Freezing the image encoder and training the text decoder helps balance performance and robustness efficiently. On the other hand, training the image encoder alone does not provide the same benefits, highlighting the text decoder's key role in maintaining robustness.

Future work will explore changes in activation functions for additional gains in adversarial robustness~\cite{sooksatra2023relu}. Current and future work supports the goals of Ethical AI by making AI systems safer and reliable. By focusing on specific parts of the model, we can make the adversarial defense process more efficient. This targeted approach can lead to the development of AI systems that are both resource-efficient and secure for public use. Enhancing the robustness of AI models against adversarial attacks builds public trust and confidence in AI systems, especially in critical applications like healthcare, autonomous driving, and content moderation.

\section{Acknowledgment}
This research work was part of a project funded by the National Science Foundation under grants CNS-2210091, CHE-1905043, and CNS-2136961.

\bibliographystyle{ACM-Reference-Format}
\bibliography{sample-base}

%%
%% If your work has an appendix, this is the place to put it.
\appendix

\end{document}